\documentclass{article}

\PassOptionsToPackage{numbers, compress}{natbib}

\usepackage[final]{nips_2017M}
\usepackage[utf8]{inputenc} 
\usepackage[T1]{fontenc}    
\usepackage{hyperref}       
\usepackage{url}            
\usepackage{booktabs}      
\usepackage{amsfonts}      
\usepackage{nicefrac}      
\usepackage{microtype}     
\usepackage{amsmath}
\usepackage{bbm} 
\usepackage{tikz}
\usepackage{multirow}
\usepackage{multicol}
\usepackage{booktabs}
\usepackage{pifont}
\usepackage{float}
\usepackage{epstopdf}
\graphicspath{{../images/}}
\DeclareGraphicsExtensions{.pdf,.jpeg,.png,.eps}
\usepackage{enumitem}
\setlist[itemize]{leftmargin=0.5cm}

\title{Bag-Level Aggregation for Multiple Instance Active Learning in Instance Classification Problems}

\author{
  Marc-André Carbonneau, Eric Granger and Ghyslain Gagnon \\
  École de technologie supérieure, Université du Québec\\
  Montreal, Canada \\
  \texttt{marcandre.carbonneau@gmail.com, \{eric.granger, ghyslain.gagnon\}@etsmtl.ca} \\
}

\begin{document}
\maketitle

\begin{abstract}
A growing number of applications, e.g. video surveillance and medical image analysis, require training recognition systems from large amounts of weakly annotated data while some targeted interactions with a domain expert are allowed to improve the training process. In such cases, active learning (AL) can reduce labeling costs for training a classifier by querying the expert to provide the labels of most informative instances. This paper focuses on AL methods for instance classification problems in multiple instance learning (MIL), where data is arranged into sets, called bags, that are weakly labeled. Most AL methods focus on single instance learning problems. These methods are not suitable for MIL problems because they cannot account for the bag structure of data. In this paper, new methods for bag-level aggregation of instance informativeness are proposed for multiple instance active learning (MIAL). The \textit{aggregated informativeness} method identifies the most informative instances based on classifier uncertainty, and queries bags incorporating the most information. The other proposed method, called \textit{cluster-based aggregative sampling}, clusters data hierarchically in the instance space. The informativeness of instances is assessed by considering bag labels, inferred instance labels, and the proportion of labels that remain to be discovered in clusters. Both proposed methods significantly outperform reference methods in extensive experiments using benchmark data from several application domains. Results indicate that using an appropriate strategy to address MIAL problems yields a significant reduction in the number of queries needed to achieve the same level of performance as single instance AL methods.
\end{abstract}

\section{Introduction}
\label{sec:introduction}

Recent years have witnessed substantial advances of machine learning techniques that promise to address many complex large-scale problems that were previously thought intractable. However, in many applications, annotating enough representative training data to train a recognition system is costly, and in such cases, one can resort to AL to reduce the annotation burden \cite{Freund1997QC,Dasgupta2011twofaces}. Moreover, several applications allow to leverage some targeted interactions with human experts, as needed, to label informative data and drive the training process. AL has been used in various applications to reduce the cost of annotations, e.g., in medical image segmentation \cite{Konyushkova2015}, text classification \cite{Tong2001,Hoi2006text} and visual object detection \cite{Vijayanarasimhan2014}. 

Alternatively, the cost of annotations can be reduced through weakly supervised learning. It generalizes many kinds of learning paradigms including semi-supervised learning and MIL in partially observable environments or learning from uncertain labels. With MIL, training instances are grouped in sets (commonly referred to as bags), and a label is only provided for an entire set, but not for each individual instance. MIL has also been shown to efficiently reduce annotation costs in several applications such as object detection (where labels are obtained for whole images) \cite{Ren2016}, description sentences \cite{Xu2016Vid,Karpathy2015,Fang2015} and web search engine results \cite{Zhu2015objDisc}. This is particularly attractive for medical image analysis where a system can learn using labeled images that were not locally annotated by experts \cite{Quellec2017}. Other successful applications of MIL include text classification \cite{Ray2005,Zhang2013MI2LS}, sentiment analysis \cite{Kotzias2015kdd}, and sound classification \cite{Briggs2012}.

This paper focuses on methods that are suitable for MIAL problems. Although several AL methods exist for single instance learning \cite{Settles2009survey}, only a handful of methods have been proposed to address MIAL problems \cite{Meessen2007,Settles2008,Zhang2010AMIL,Melendez2016AMIL}. Single instance active learning (SIAL) methods are not suitable for MIL because: 1) in MIL, instances are grouped in sets or bags, and 2) training instances have weak labels. The arrangement of instances into bags gives rise to several different tasks, such as bag classification and instance classification which must be addressed differently \cite{Carbonneau2016Survey}.  

Different learning scenarios exist for active MIL \cite{Settles2008}. In this paper, we focus on the scenario where the learner has a set of labeled bags at its disposal, and must predict the label of each individual instance. The learner can query the oracle to label the bag's content. The final objective is to uncover the true labels of the instances, which corresponds to the transduction setting described in \cite{Garcia2011degrees}. Given instances that are correctly labeled, any classifier can be used in a supervised fashion to classify instances not belonging to the training set in an inductive setting \cite{Garcia2011degrees}. To our knowledge, this scenario has never been studied in the literature. The few existing MIAL methods focus on bag classification \cite{Meessen2007,Settles2008,Zhang2010AMIL} or select groups of instances in a scenario where there is only one query round \cite{Melendez2016AMIL}. 

The MIAL scenario that we address is relevant in several real-world problems. For example, in some computer-assisted diagnosis applications, classifier is trained to identify localized regions of organs or tissues afflicted by a given pathology. A classifier is typically trained using afflicted regions identified by an expert or a committee of experts, which is costly in terms of time and resources. This limits the quantity of available data for training. However, it is easier to obtain images along with a subject diagnosis as a weak label (bag label). In order to make better use of the experts, the MIAL learner identifies the subject whose local annotations would most improve the classifier. In this example, we believe that our learning scenario is more plausible than the second scenario where instances are queried individually. When experts are asked to provide local annotations of afflicted tissues or organs, it makes more sense to provide an entire image (bag) of the patient rather than provide isolated regions (instances). In this kind of applications, it is important for the annotator to be aware of the context provided by the surroundings of the segment when assigning a label. A similar argument can be made for text classification where an instance can be a sentence or a paragraph. It is easier to provide an accurate label for individual parts with knowledge of the entire text.

Beyond the well-known difficulties associated with AL, MIL instance classification raise several challenges. First, leveraging the weak supervision provided by bag labels is challenging because it is not explicitly known how each instance relates to its bag label. Also, the fact that training instances are arranged in sets adds an extra layer of complexity regarding relations between training instances. Moreover, in MIL, instance classification is often associated with severe class imbalance problems. Finally, AL and weakly supervised learning are often used to reduce the annotation cost of large amount of data which calls for algorithms with low computational complexity. For cost-effective design of an instance classifier through MIL, an AL algorithm should:
\begin{itemize}
\item characterize uncertainty in the instance space -- assess which regions of the instance space are	most ambiguous to the classifier, and thus informative for design.
\item identify the most informative bag for the learner given multiple regions of the instance space. 
\item leverage bag label information, from queried and non-queried bags. This is in contrast to traditional AL problems because in our context bag labels provide weak indication of the instance labels.
\end{itemize}

Two new MIAL methods are proposed in this paper for bag-level aggregation of instance informativeness, allowing to  select the most informative bags to query, and then learn. The first method -- \textit{aggregated informativeness} (AGIN) -- assesses the informativeness of each instance to compute the informativeness of bags. Informativeness is based on classifier uncertainty, and instances near the decision boundaries are prioritized. The second method -- \textit{cluster-based aggregative sampling} (C-BAS) -- characterizes clusters in the instance space by computing a criterion based on how much is known about the cluster composition and the level of conflict between bag and instance labels. The criterion enforces the exploration of the instance space and promotes queries in regions near the decision boundary. Moreover, the criterion discourages the learner from querying about instances for which the label can be inferred from bag labels. Extensive experiments have been conducted to assess the benefits of using both proposed methods in three application domains: text, image and sound classification. 

The rest of the paper is organized as follows. The next section reviews the state-of-the-art in active MIL. Section \ref{Section:ProposedMethods} formalizes the active MIL problem and presents the two proposed methods. The experimental methodology is described in Section \ref{Section:Experiments}, and results are analyzed and discussed in \ref{Section:Results}.

\section{Multiple Instance Active Learning}

This paper focuses on pool-based AL methods \cite{Settles2009survey} where the learner is supplied with a collection of unlabeled and labeled samples. The learner must select the best instance, or groups of instances, to query. Pool-based AL problems have been tackled following two intuitions \cite{Dasgupta2011twofaces}: 1) queried instances should shrink the classifier hypothesis space as much as possible, and 2) cluster structure of the data should be exploited for efficient exploration of the input space. The methods proposed in this paper address the MIAL problem from each intuition perspective.
 
Several types of approaches shrink the classifier hypothesis space. The methods based on uncertainty query the most ambiguous instances for the classifier \cite{Tong2001,Lewis1994} or the instance causing the most disagreement in a pool of classifiers \cite{Seung1992,Melville2004}. A drawback of these methods is that they tend to choose outliers for query since they are often ambiguous for the classifier \cite{Tang2002,Zhu2008}. To avoid this problem, some methods compute the expected error reduction \cite{Roy2001,Guo2007} or expected model change \cite{Settles2008}. They estimate the impact of obtaining each individual instance label on the generalization error or the model parameters. However, these methods are computationally expensive because classifiers must be trained for each possible label assignment of each unlabeled data sample. To avoid this problem, some methods aim to reduce generalization error by minimizing the model variance \cite{Hoi2006text,Cohn1994ALstat}, typically by inverting a Fisher information matrix for each training instance. The size of the matrix depends on the number of parameters in the model which can rapidly become intractable \cite{Settles2009survey}. All these approaches are subject to sampling bias problems \cite{Dasgupta2011twofaces}, where some true instance labels may never be discovered for multi-modal distributions. This is because at the start of the learning process a classifier is trained using sampled data, and then later, queries are proposed near the decision boundaries of this classifier. If data structure exists, but was not captured by the initial samples, it may never be discovered.

Another group of AL methods relies on the characterization of the data distribution in the input space \cite{Settles2008ID,Fujii1998,Nguyen2004}. Instead of concentrating on decision boundaries, they assess the structure of input data in order to query for informative instances that are representative of the input distribution. Leveraging the input data structure promotes exploration and discourages the selection of outliers. As a result, methods characterizing the input space yield better performance than other types of method when the quantity of labeled data is limited. However, as more labels are queried, methods that shrink the hypothesis space tend to perform better \cite{Wang2015QDR}. The complexity of these approaches is generally similar to other kind of approaches with an added initial cost of a clustering or density estimation step \cite{Settles2008ID}. 

As will be described in Section \ref{Section:ProposedMethods}, the AL methods proposed in this paper follow these two different intuitions. AGIN seeks to shrink the hypothesis space based on classifier uncertainty, while C-BAS characterizes the data distribution. These methods have been developed with computational efficiency in mind, which is increasingly important to address the growing complexity of large-scaled applications.

Although MIL methods were initially proposed for bag classification \cite{Amores2013}, instance classification problems have more recently attracted growing interest \cite{Xu2016Vid,Zhu2015objDisc,Vanwinckelen2015,Vezhnevets2010}. These  are different tasks that require different approaches \cite{Carbonneau2016Survey,Vanwinckelen2015}. MIL methods fall into one of two main categories depending on which level, bag or instance, discriminant information is extracted \cite{Amores2013}. Bag-level methods compare bags directly using set distance metrics or embed bags in a single summarizing feature vector \cite{Chen2006,Wang2000citation,Cheplygina2015DBE,Gartner2002,Zhou2009migraph}. These methods do not perform instance classification and are unsuitable in our context. In contrast, instance-level methods predict the class of instances and combine these predictions to infer the bag label (e.g., APR \cite{Dietterich1997}, DD and EM-DD \cite{Maron1998,Zhang2001}, mi-SVM and MI-SVM \cite{Andrews02}, RSIS \cite{Carbonneau2016RSIS} and MI-Boost \cite{Babenko2008alignement}). While these methods are usually designed for bag classification, they can be employed  for instance classification tasks. It has been shown that bag classification and instance classification tasks have different misclassification costs \cite{Carbonneau2016Survey}, which means that the best bag classifier is not necessarily the best instance classifier \cite{Vanwinckelen2015}. Moreover, experiments in \cite{Ray2005,Carbonneau2016Survey} show that single instance classifiers often perform comparably to MIL methods, especially for instance classification.

The literature on MIAL is limited and almost each method is proposed for a specific learning scenario. There are methods that query bag labels for bag classification. The method in \cite{Meessen2007} embeds bags in a single feature vector using a representation based on MILES \cite{Chen2006}. An SVM is used for classification and the embedded bags which are closest to the decision hyper-plane are selected as in \cite{Tong2001}. This method has been generalized in \cite{Zhang2010AMIL} and a selection method based on Fisher's Information criterion has also been proposed. The learning scenario in \cite{Settles2008} is similar to ours in that all bag labels are known and the learner queries instance labels from positive bags. However, our goal is to train an instance classifier (not a bag classifier), and the learner queries all instance labels from a bag (instead of only one), which we believe to be more efficient in practice. They train a logistic regression classifier optimized for bag-level classification accuracy. Their selection method is based on uncertainty sampling and expected gradient length. Queried instances are duplicated and added to the training set as singleton bags. While this method works well in practice, it is computationally expensive and the expected gradient length method is sensitive to feature scale \cite{Settles2009survey}. The method proposed in \cite{Melendez2016AMIL} targets the instance classification task in a peculiar MIAL scenario where there is only one query round. First, instances are classified using a MIL algorithm \cite{Melendez2014novel} and then, the most valuable instances are grouped in regions. These hundreds of regions are then labeled by an expert and the MIL classifier is retrained. This differs from the scenario in this paper because there is only one query round, and the expert must annotate a region instead of an image.

\section{Proposed Methods}
\label{Section:ProposedMethods}

\begin{figure*}[!t]
\centering
\includegraphics[width=1.0\textwidth]{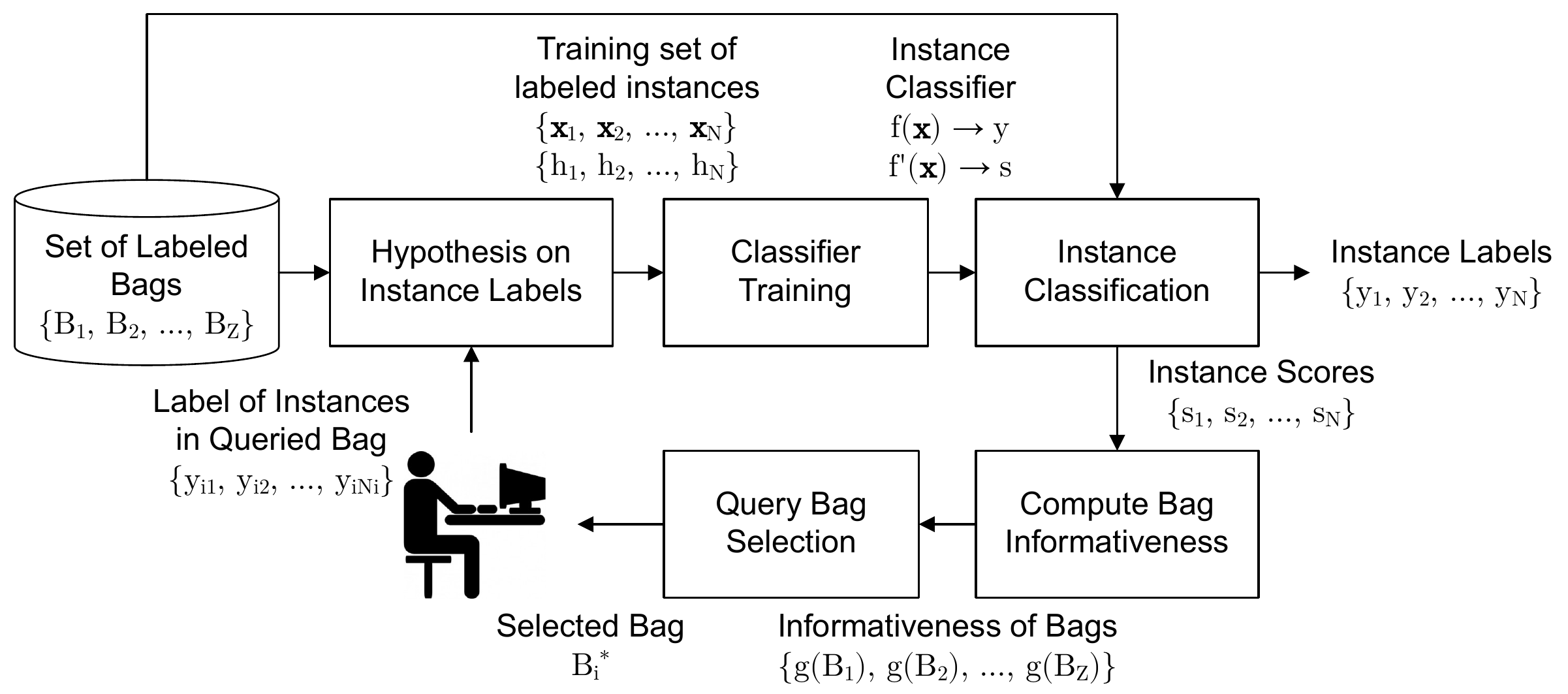}
\caption{Block diagram of the general operations performed in our MIAL scenario for instance classification. The learner is initially supplied with a set of labeled bags, but no instance label. During each iteration, the learner predicts a label for each instance. An instance classifier is then trained, and used to assign a label and a score to all instances in the training set. The score of each instance is used to identify the most informative bag to query. Finally, the labels of all instances in the selected bag are annotated by the oracle in order to update the hypothesis and retrain the classifier.}
\label{Fig:TOP}
\end{figure*}

Figure \ref{Fig:TOP} presents an overview of the MIAL framework for our learning scenario. The training data set $\mathcal{B} = \{ B_1, B_2, ..., B_Z \}$ is a set of $Z$ bags, each one is associated with a label $Y_i \in \{-1,+1\}$ and contains $N_i$ instances: $B_i = \{\textbf{x}_{i1}, \textbf{x}_{i2}, ..., \textbf{x}_{iN_i}, \}$. Each instance $\textbf{x}_{ij}$ has an associated label $y_{ij}\in \{-1,+1\}$. All the bag labels are known a priori. Following the standard MIL assumption \cite{Dietterich1997}, the labels of instances in negative bags are assumed to be negative, while positive bags contain negative instances and at least one positive instance:

\begin{equation}
Y_i = \left\{\begin{matrix}
+1 & \text{if} & \exists y \in B_i:\; y_{ij}=+1; \\ 
-1 & \text{if} & \forall y \in B_i:\; y_{ij}=-1.
\end{matrix}\right. 
\end{equation}  

The task consists in training a classifier to correctly predict the label of each individual instance $f(\textbf{x}) \rightarrow y$. The classifier's decision function can be iteratively improved by querying an oracle about a bag. To select the most informative bag for query, the function $g(B) \rightarrow \mathbb{R}_{\geq0}$ assigns an informativeness score to each of them. Once a bag has been selected for query ($B^*$), the oracle provides labels for all its instances. Then, the hypothesis on instance labels $h_{ij}$ is updated, and the classifier is retrained. The next best candidate bag for query is selected, and so on. The rest of this section presents two new methods to derive $g(B)$ for selecting bags for query.

\subsection{Aggregated Informativeness (AGIN)}
\label{Section:AGIN}
This method is inspired from SIAL methods (like in \cite{Tong2001}) that select the instance expected to provide the largest reduction in the set of all consistent hypotheses. For instance, when working with SVM classifiers, this amounts to selecting the instance which is the closest to the decision hyper-plane. However, in MIL problems, instances are grouped into bags and the bag containing the single most informative instance is not necessarily the optimal choice. If the most informative instance is part of a bag containing only trivial instances, it may be advantageous to select another bag containing several difficult instances, even if none of them are the single most informative instance in the entire data set. In other words, a bag should be selected based on the combined informativeness of its instances. 

Here we describe the method as an adaptation of \cite{Tong2001}. The SVM classifier is used as an example, but it can easily be replaced with any type of classifier. First, the distance to the decision hyper-plane must be transformed into instance informativeness. Let $f'(\textbf{x}) \rightarrow s$ be a function returning a classification score $s \in \mathbb{R}$ for an instance $\textbf{x}$. This is the same as the classifier function $f(\textbf{x})$, without a decision threshold.

For an SVM, the decision hyper-plane is defined by  $f'(\textbf{x}) = 0$. The informativeness of an instance can be obtained using a radial basis function $\phi(\textbf{x})$ centered at 0. Any type of function can be used as long as it is maximized at the decision threshold, and it decreases monotonically with distance. In this paper we use: 
\begin{equation}
\phi(\textbf{x}) = e^{-2\left | f'(\textbf{x})  \right |} 
\label{Eq:scoreInformativeness}
\end{equation}
This function  decreases exponentially as the magnitude of $s$ increases. This ensures that instances located close to the hyper-plane are highly prioritized over other less ambiguous instances. 

The informativeness score of a bag is the aggregation of informativeness scores over all its instances:
\begin{equation}
g(B) = \sum_{\textbf{x} \in B} \phi \left( \textbf{x}  \right )
\label{Equation:Agrregation}
\end{equation}
The bag ($B^*$) with the highest informativeness score is selected for query:
\begin{equation}
B^*=\underset{B \in \mathcal{B}}{\text{argmax}} \; g(B)
\label{Equation:argmax}
\end{equation}

\subsection{Clustering-Based Aggregative Sampling (C-BAS)}
\begin{figure*}[!t]
\centering
\includegraphics[width=1.0\textwidth]{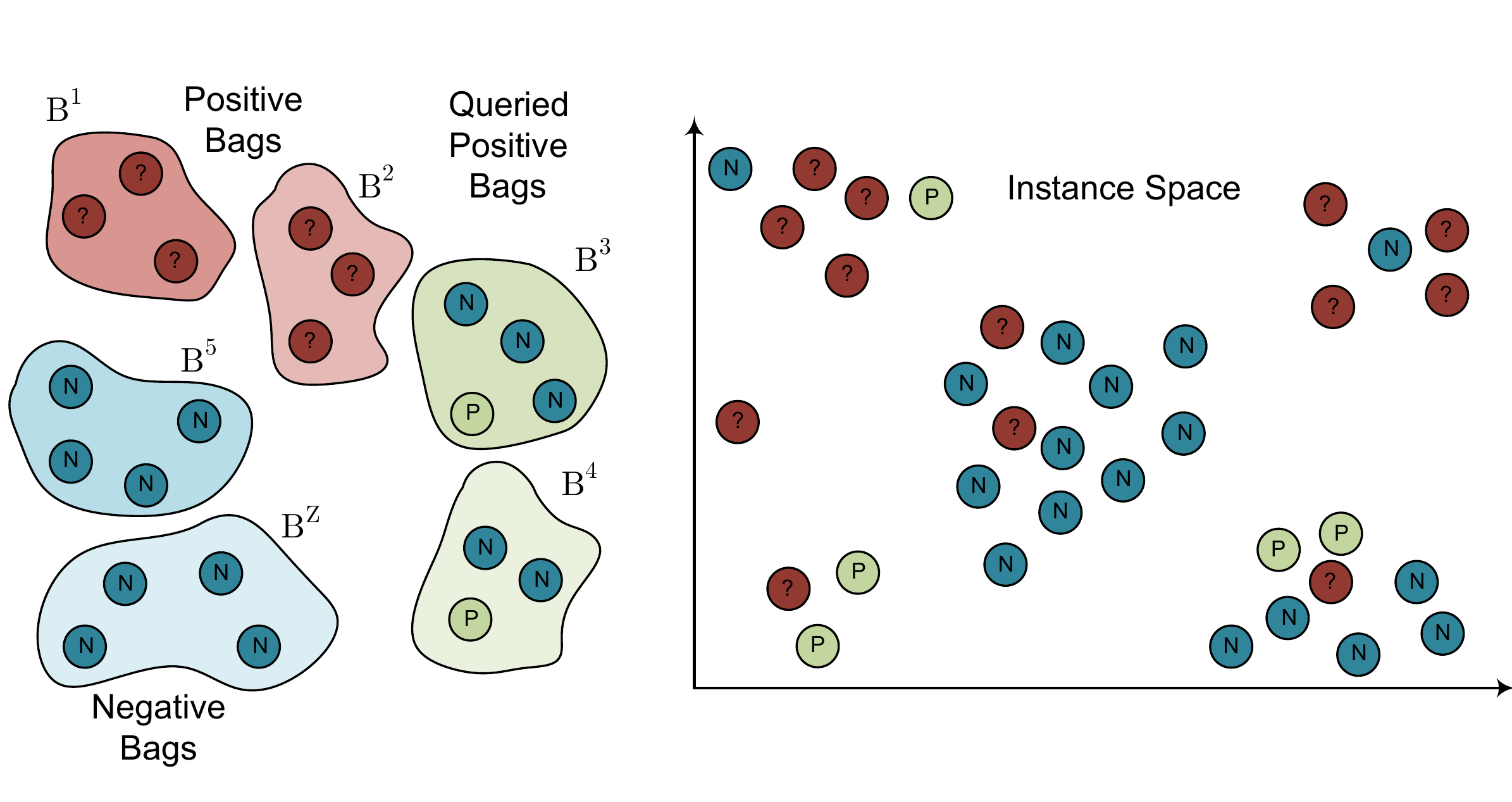}
\caption{Representation of clusters in the instance space in an MIAL problem. It shows different types of cluster. In cluster a), even if none of the instance have been queried, they are considered non-informative because they all belong to bags of the same class. The same can be said about instances in cluster b). In cluster c) and d), all labeled instances belong to the same class even if their bag labels are different. The remaining instances are therefore deemed to be uninformative. Most of the instance labels in cluster e) are known and thus, the label of the remaining instance is unlikely to provide useful information. Instances in cluster f) should be informative because there is label disagreement at bag and instance level, and an appreciable proportion of instance labels remain to be discovered.}
\label{Fig:ClusterTypes}
\end{figure*}

This method is proposed to alleviate problems associated with the sample bias, and to leverage the weak information provided by bag labels and classifier predictions on instance labels. The intuition behind C-BAS is that a cluster of instances should meet three conditions to be informative: 1) bag label disagreement, 2) instance label disagreement, and 3) contain a considerable proportion of non-queried labels. If a cluster contains instances from only one class of bags, the label of these instances is the same as the label of their bag. Obtaining the true labels for these instances is not informative. Inversely, if a cluster contains different types of instances, it should define a decision boundary. Acquiring labels in this cluster is likely to help refine the overall decision boundary. Finally, to encourage exploration, clusters for which few labels are known will be considered as informative. Figure \ref{Fig:ClusterTypes} illustrates these situations.  

C-BAS starts by hierarchical clustering of data in the instance space. As in \cite{Dasgupta2008}, we employ agglomerative hierarchical clustering, although it can be replaced with any type of hierarchical clustering algorithm. This type of method does not require setting the number of expected clusters a priori, and creates a clustering dendogram or tree that is used to create space partitioning of different granularities. The informativeness of instances in each cluster $k$ is evaluated by a criterion $c_k$ that accounts for cluster composition of the cluster. The criterion is composed of 3 terms enforcing the aforementioned conditions of informativeness:
\begin{equation}
c_k = BD_k \cdot ID_k \cdot E_k 
\end{equation} 
The $BD_k$ term measures the level of disagreement between bag labels with an entropy-based function:
\begin{equation}
BD_k = \frac{\beta \textrm{log}(\beta) +  (1-\beta)\textrm{log}(1-\beta)}{\textrm{log}(0.5)},
\end{equation} 
where $\beta$ is the proportion of instances from positive bags among the instances assigned to the cluster. If all instances come from bags of the same class, this term is equal to 0 which inhibits further research in this cluster. When bag labels are equally divided among the two classes, the term value is equal to 1.  Similarly, the $ID_k$ term measures the degree of disagreement between instance labels:
\begin{equation}
ID_k = \frac{\zeta \textrm{log}(\zeta) +  (1-\zeta)\textrm{log}(1-\zeta)}{\textrm{log}(0.5)},
\end{equation}
where $\zeta$ is the proportion of positive instances among the instances assigned to the cluster. When the true label of an instance remains unknown, the classifier's prediction is used as label. Finally, The term $E$ promotes cluster exploration based on the proportion of unlabeled instances ($\alpha$) in contains:
\begin{equation}
E_k = \frac{1-e^{-\alpha}}{1-e^{-1}},
\end{equation}
When all instance labels are known this terms is equal to 0, and when none are known, it is 1 .

\subsubsection*{Exploring the Clustering Tree}
The clustering tree is explored from top to bottom. Iteratively the tree is pruned farther away from the trunk, each time yielding a clustering of finer granularity. For each clustering level $l \in \mathcal{L}$, the informativeness criterion $c_k$ of each cluster $k$ is computed. The informativeness $\phi(\textbf{x})$ of an instance is an accumulation of the informativeness of each cluster $k$ to which it was assigned:
\begin{equation}
\phi (\textbf{x})=\sum_{l \in \mathcal{L}} \sum_{k \in \mathcal{K}_l}  \mathbbm{1}_k (\textbf{x}) \cdot c_k,
\end{equation}  
where $\mathcal{K}_l$ is the set of clusters obtained when the tree is cut at level $l$. 

Different levels of granularity are necessary to correctly assess the informativeness of instances. By considering only large clusters obtained (top of the tree), all instances would be provide the same level of information. They would all be assigned to few large clusters which are likely to present a high level of disagreement between labels, and include many non-queried instances. Inversely, by considering very fine cluster granularity (bottom of the tree), the levels of disagreement between labels $BD_k$ and $ID_k$ tend towards 0, which means $c_k = 0$ and thus $\phi (\textbf{x})=0$ for all $\textbf{x}$. This is equivalent to randomly picking any unlabeled instances. Accumulating evidences on informativeness over levels of cluster granularity allows to compromise between the two extreme cases. Once all instance informativeness scores $\phi(\textbf{x})$ are computed, the query bag $B^*$ is selected in the same way as for AGIN (see (\ref{Equation:Agrregation}) and (\ref{Equation:argmax})).

\section{Experiments}
\label{Section:Experiments}

\begin{table*}[!ht]
\caption{Summary of the properties of the benchmark data sets.}
\label{Table:dataDesc}
\centering
\begin{tabular}{lcccccccccc}
\toprule
    	&	&  	&   &  	&\multicolumn{3}{c}{Inst. per Bag} &\multicolumn{3}{c}{Class imbalance} \\
\cmidrule(r){6-8} 
\cmidrule(r){9-11}
Name & Sets & Bags	& Inst.	& Feat.		& Min 	& Max 	& Avg & Min 	& Max 	& Avg\\
\hline
SIVAL \cite{Settles2008,Rahmani2005}	&25	& 180 & 5690 & 30 & 31 & 32 & 32 & 0.035 & 0.218	& 0.095 \\
Birds \cite{Briggs2012} &13 & 548	& 10232 & 38 & 2 & 43 & 19 & 0.003 & 0.143 & 0.040	\\
Newsgroups \cite{Settles2008}&20	& 100	& 4060 & 200 & 8 & 84	& 40 & 0.012 & 0.035	& 0.018	\\
\bottomrule
\end{tabular}
\end{table*}

All experiments were repeated 100 times and conducted with the following protocol. The data sets were randomly split in test (1/3) and training (2/3) subsets. For fair comparison, all MIAL methods are the same except for the bag selection scheme. The initial hypothesis for the labels individual instance is that they inherit the label of their bag, which is often successful in practice \cite{Ray2005,Carbonneau2016Survey}. Bags are queried one by one until there are no positive bags left to query in the training set. After each query, the performance of classifiers is measured on the training and test subsets. This corresponds to the transductive and inductive learning settings described in \cite{Garcia2011degrees}.

As bags are queried, class imbalance of instance labels grows, which is an important concern for MIL instance classification tasks \cite{Herrera2016imbalance}. This is particularly true in data sets where the proportion of positive instances in positive bags is low. We handle class imbalance using Different Error Costs SVM (DEC-SVM) \cite{Veropoulos1999}. This SVM method assigns different misclassification costs $C$ to different classes. Table \ref{Table:SVMParameters} reports the configuration of the SVM used for each data set. These parameters were obtained with 5-fold cross-validation using the real instance labels. We used the LIBSVM implementation \cite{LIBSVM}. The ratio between the misclassification penalty cost of the classes corresponds to the class imbalance ratio ($\rho=\nicefrac{N_+}{N_-}$). $N_+$ and $N_-$ are the number of positive and negative instances in the training set. Each time an SVM is trained, class imbalance ratio is recomputed and misclassification costs are adjusted accordingly.

Performance is reported in terms of $F_1$-Score and the area under the precision-recall curve ($AUC_{PR}$) which are appropriate metrics for problems with class imbalance.

\begin{table}[!ht]
\small
\centering
\caption{SVM parameter configuration used in experiments} 
\begin{tabular}{lcccc}
\toprule

\textbf{Dataset}& $C_+$& $C_-$ & kernel & 
$\gamma$ \\

\midrule
SIVAL 		& 1000 & $\rho$1000&Gaussian RBF & 0.01 \\
Birds 		& 1000 & $\rho$1000&Gaussian RBF & 0.1 \\
Newsgroups 	& 1000 & $\rho$1000& $\chi^2$ & - \\
\bottomrule
\end{tabular}
\label{Table:SVMParameters}
\end{table}

To assess the benefits of employing bag selection schemes for query selection, the first reference method selects bags at random. It selects only positive bags since the label of instances in negative bags are assumed to be known. The few MIAL methods proposed in literature were not designed for instance classification, so the simple margin method \cite{Tong2001} was considered as the second reference method. It consists in picking the closest unlabeled instance to the decision hyper-plane of the SVM. In our experiments the method selects the bag containing this most informative instance. This method is originally intended for single instance learning scenarios and is closely related to AGIN. It is therefore relevant to show the effect of the proposed aggregation schemes.

\subsection{Data Sets}
The MIAL methods are evaluated using the three most widely used collection of MIL data sets providing instance annotations: Birds \cite{Briggs2012}, SIVAL and Newsgroups. The last two were introduced to compare MIAL methods in \cite{Settles2008}. They represent 3 different application domains -- content-based image retrieval, text and sound classification. Each dataset contains different classes which are in turn used as the positive class yielding a total of 58 different problems. Table \ref{Table:dataDesc} gives an overview of the properties for each data set. 

\begin{figure*}[t!]
\centering
\includegraphics[width=1.0\textwidth]{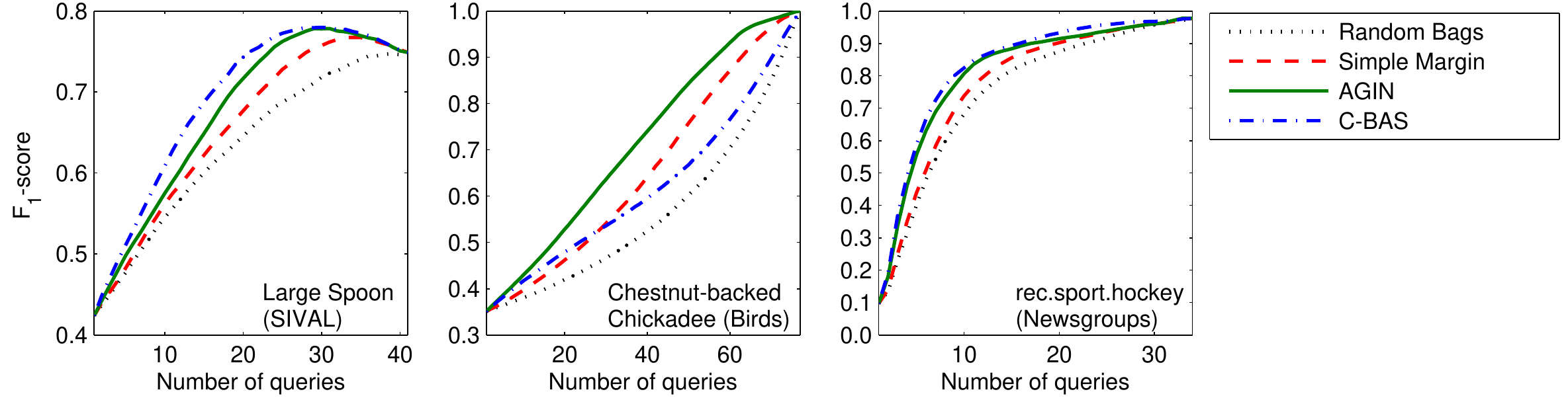}
\caption{Average learning curves for MIAL methods on SIVAL, Birds and Newsgroups datasets.}
\label{Fig:GraphTrain}
\end{figure*}

\subsubsection{SIVAL} 
The Spatially Independent, Variable Area and Lighting (SIVAL) data set for visual object retrieval \cite{Rahmani2005} contains 1500 images each depicting one of 25 complex objects photographed from different viewpoints in various environments. The version used in this paper has been segmented and hand-labeled to compare MIAL approaches in \cite{Settles2008}. Each object is in turn considered as the positive class, and all remaining objects are part of the negative class. This yields 25 different 2-class learning problems. Each of the 25 data sets contains 60 positive images and 120 negative images sampled uniformly from all 24 negative classes. Images are represented as bags which are a collection of segments. Texture and color features are extracted from segments as well as neighborhood information yielding a 30-dimensional feature vector for each. The proportion of positive instances in positive bags is 25.5\% in average and ranges from 3.1\% to 90.6\%. This data set exhibits high intra-class variation which means that the positive instance distribution is multimodal. 

\subsubsection{Birds}
This data set \cite{Briggs2012} contains recordings of bird songs captured by unattended microphones in the forest. Each bag is the spectrogram of a 10 seconds recording. The recording is temporally segmented and 38 features characterizing shape, time and frequency profile statistics are extracted from each segment. The data set contains 13 species of birds, which are in turn considered as the positive class yielding 13 problems. This data set is difficult because in some cases there is extreme class imbalance at bag and instance level. For example, there are only 32 instances out of 10232 that belong to the hermit thrush. In the best case, positive instances represent 12.5\% of all instances. As opposed to the other data sets, each class (except for background noise) is represented by a single compact cluster in space.

\subsubsection{Newsgroups}
This MIL data set was created using instances from the \textit{20 Newsgroups} data set corpus in \cite{Settles2008}. Instances are posts from newsgroups about 20 different subjects. Each post is represented by a 200 term frequency-inverse document frequency feature vector. For each version of the data set, a subject is selected as the positive class and the remaining 19 other subjects constitute the negative class. A bag is a collection of posts. The feature vectors used for this data set are sparse histograms which makes the distribution different from the two other problems. It constitutes a good way to evaluate the robustness of the proposed method to different data distribution types. Moreover, the average proportion of positive instances in positive bags is rather low, which also makes the problem difficult and accentuate problems related to class imbalance.

\subsection{Implementation Details for C-BAS}
\label{Section:DetailsCBASS}
Here we detail the particular implementation of C-BAS that we use in the experiments. The clustering three is obtained using the Ward's average linkage algorithm. We then obtain different clustering refinements by cutting the tree at different levels. To make sure to cut at significant levels in the tree, we compute the inconsistency coefficient $\delta$ of all links in the tree: 
\begin{equation}
\delta_k = \frac{h_k-\mu_{\mathcal{N}_k}}{\sigma_{\mathcal{N}_k}},
\end{equation}
where $h_k$ is the height of the link $k$ (cophenetic distance between the clusters). The set $\mathcal{N}_k$ contains all links in the $P$ hierarchical levels under $k$. $\mu_{\mathcal{N}_k}$ and $\sigma_{\mathcal{N}_k}$ are the average and the standard deviation of the height of the links contained in $\mathcal{N}_k$. A high inconsistency coefficient means that the two clusters joined by the link are farther apart then the clusters linked in the levels below, which indicates a natural separation in the data structure. 

Once the inconsistency coefficients $\delta$ has been computed for all links, they are sorted from highest to lowest. Clusters are obtained using these values as thresholds. Instances or clusters can only be linked together if the inconsistency coefficient of the link is lower than the threshold. Iteratively, the threshold is lowered and finer clusterings are obtained. In the experiments of this paper, we use 20 threshold levels and $P$ has been arbitrarily set to 16 for all data sets. Both parameters could be optimized depending on the application.

\section{Results and Discussion}
\label{Section:Results}

MIAL methods are evaluated based on their ability to uncover the true instance labels in the training set (transductive learning task) and to classify a test set with a classifier trained using these uncovered labels (inductive learning task). Fig. \ref{Fig:GraphTrain} shows an example (over 100 runs) of the evolution of average $F_1$-score values on the training subset as a function of the number queries to the oracle. Similar learning curves were obtained with $AUC_{PR}$ but are not shown here since they do not provide pertinent additional information. Results show that for each data set, the proposed methods can significantly improve the learning process. Each curve starts (no bags have been queried) and finishes (all true instance labels are known) at the same level of performance.  

From these curves, it is possible to see how many queries are necessary to achieve the same level of performance with different methods. For example, selecting random bag may necessitate as much as 23 (out of 40) more queries than C-BASS to obtain the same $F_1$-Score on the Glaze Wood Pot training set. This is a best case scenario but nonetheless, out of the 58 data sets, using AGIN has lead to a reduction of the number of query necessary on all but 1 test data set with the $AUC_{PR}$ metric. Similarly, C-BASS has resulted in a query reduction for all but 2 data sets.

In some of these curves, after a certain number of queries, the performance starts to decrease (see Fig. \ref{Fig:GraphTrain}). While it seems counter-intuitive, this can be explained by the fact that the metric reported in the graph is different from the surrogate loss function used as an optimization objective. In our case, the SVM optimizes the hinge loss over all instances which does not guarantees the optimization of the $F_1$-Score (see \cite{Loog2012Dipping,Loog2017Meas} for a more detailed discussion on the subject). 

To compare the overall performance of methods for the entire AL sequence, the normalized area under the learning curve (NAULC) was used for both $F_1$-score and $AUC_{PR}$ metrics. It corresponds to the area under curves as displayed in Fig. \ref{Fig:GraphTrain} divided by the total number of queries. For each problem in each data set, we compute the average NAULC and identify the best performing method as a win. Statistical significance of results is assessed using a t-test ($\alpha$=5\%). Table \ref{Table:Training} reports the number of wins for all methods (complete result tables can be found in the supplementary material document). Both proposed methods outperform the reference methods for all three application domains and for both the transductive and inductive tasks. Results indicate that aggregating instance informativeness to select queried bags is a better strategy than selecting the most ambiguous instance, and that SIAL methods should be adapted to MIL problems to improve performance.

Results suggest that proposed methods are better suited for different type of data. For example, AGIN outperforms other methods on the Birds dataset, while C-BAS yields better results with SIVAL data. Indeed, the positive instances in Birds data are likely to be grouped in very few clusters since birds of the same specie tend to have similar songs. In that case, the best strategy is to concentrate on refining the decision boundary since there are no hidden cluster structure to discover. Inversely, the positive distribution in SIVAL data is likely to have several modes. The appearance of an object, and thus its corresponding feature representation, can be very different depending on point-of-view, scale and illumination. In that case, it is important to discover these multiple clusters as rapidly as possible, which favors the C-BAS approach.

\begin{table*}[!ht]
\caption{Number of wins for each algorithm on each corpus. The NAULC for 100 runs were averaged and a t-test was performed to determine the best algorithm ($\alpha = 0.05$).} 
\centering
\scalebox{0.93}{
\begin{tabular}{llcccccccc}
\toprule
&& \multicolumn{2}{c}{Random Bags}& \multicolumn{2}{c}{Simple Margin}& \multicolumn{2}{c}{AGIN}& \multicolumn{2}{c}{C-BAS}\\
\cmidrule(r){3-4} \cmidrule(r){5-6} \cmidrule(r){7-8} \cmidrule(r){9-10} 
\textbf{Task Setting}& \textbf{Dataset}& $F_1$ & $AUC_{PR}$& $F_1$ & $AUC_{PR}$& $F_1$ & $AUC_{PR}$& $F_1$ & $AUC_{PR}$\\
\hline
Transductive 		&
SIVAL 				&0 	&0 	&2 	&3 	&14 &7 	&\textbf{19}	&\textbf{23}\\
(Training set)&Birds 	&0 	&1 	&0 	&1 	&\textbf{13}		&\textbf{12}	&2 	&8\\
&Newsgroups 		&0 	&0 	&1 	&2 	&8 	&16	&\textbf{19}	&\textbf{17}\\
&TOTAL WINS 		&0 	&1 	&3 	&6 	&35 &35 &\textbf{40} 	&\textbf{48}\\
\hline
Inductive 			& 
SIVAL 				&3 	&1 	&13 &12	&\textbf{21} &\textbf{20}	&18		&19\\
(Test set)&Birds 	&2 	&5 	&3 	&4 	&\textbf{13} &\textbf{12}	&8 		&\textbf{12}\\
&Newsgroups 		&6 	&6 	&10	&17	&\textbf{20} &\textbf{20}	&18		&16\\
&TOTAL WINS 		&11 &12	&26 &33 &\textbf{54} &\textbf{52} 	&44 	&47\\
\bottomrule

\end{tabular}}
\label{Table:Training}
\end{table*}

The results in Table \ref{Table:Training} suggest that AGIN and C-BAS are better suited for different tasks. This is because uncovering the labels of instances in labeled bags is slightly different than training a classifier that generalizes well to unseen data. This has to do with how the algorithms approach the problem, class imbalance and the initial hypothesis on instance labels. The initial hypothesis that all instances inherit their bag labels ensures that all positive instances are used for training the classifier. At the same time, many negative instances are falsely labeled positive (FP). These noisy labels do not necessarily pose a serious difficulty when training the classifier. In regions densely populated with negative instances, FP are outweighed by true negative instances, and thus, overlooked by the classifier. In regions where there is a mix of true positives and negatives, FPs artificially expand the classifier positive regions which has the effect of increasing the sensitivity of the classifier. This means that, as bags are queried, precision increases but recall decreases. The initial increased sensitivity of the classifier has a beneficial effect on generalization (under these metrics) in context where there is class imbalance. Therefore preserving this effect while refining the decision boundary insures better generalization while learning. This explains why AGIN performs better for test set classification. Inversely, C-BAS uncover FP in all regions of the instance space which helps in yielding better results for the transductive task but mitigates the beneficial effect of the temporary increased sensitivity when compared to AGIN.

\begin{figure*}[h!]
\centering
\includegraphics[width=1.0\textwidth]{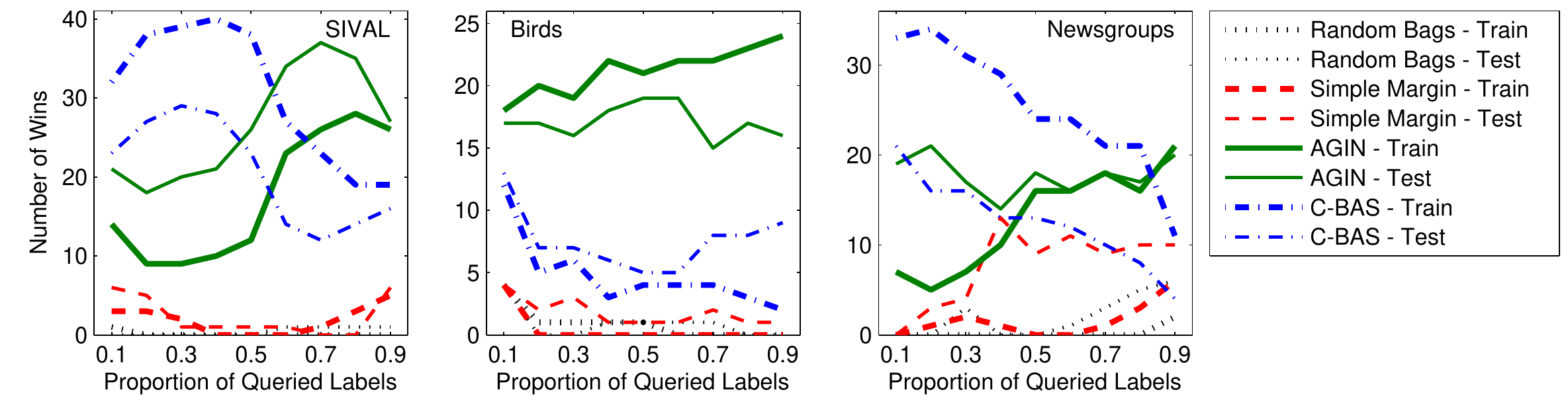}
\caption{The number of wins of each method (both metrics) vs. the proportion of queried bag labels.}
\label{Fig:EvoWins}
\end{figure*}

It had been previously shown that when very few instances are labeled, methods characterizing the distribution of the input space, like C-BAS, perform better than methods reducing the classifier hypothesis space, like AGIN, and vice-versa \cite{Wang2015QDR}. This is observed in our experiments (see Fig. \ref{Fig:EvoWins}). This is because C-BAS pushes the learner to quickly explore the most promising data clusters through the $E$ term. Moreover, the $BD$ term prevents the learner from querying instance labels that can be inferred from bag labels. After a certain number of queries, it becomes more important to refine decision boundaries, and that is when AGIN performs better. 

For instance classification problems in MIAL, the exploration of the instance space is always promoted indirectly, which reduces the severity of sample-bias problems as found in SIAL problems. This implicit exploration comes from the fact that all instances of a queried bag are labeled together. Even if a bag is selected because it contains instances near a decision boundary, the other instances in the bag provide information about other regions of the instance space. This helps AGIN achieve a high level of performance.

Based on these experiments, it seems that the AGIN method is preferable to the others in many situations. It achieves a high level of accuracy while remaining fairly simple to implement. It exhibits competitive levels of performance in both transductive and inductive learning tasks. There are two situations where it is preferable to use C-BAS: 1) when there are few known labels, and 2) when the positive instances are distributed in several regions of the input space. 

While the proposed methods perform well with the type of data used in our experiments, we believe that there are some types of MIL problems were they might not yield optimal performance. As explained in \cite{Carbonneau2016Survey} MIL problems can possess several characteristics which require special care. Some of them would probably be difficult to address with the proposed algorithms. For example the proposed methods assume that all features are relevant for classification. This makes it difficult to deal with MIL data presenting strong intra-bag similarity. This means that instances from the same bag are similar and thus located in the same region of space.  Also, AGIN and C-BASS were developed under the standard MIL assumption where all instances in negative bags are assumed to be negative. This assumption is sometimes violated in practice. Finally, the algorithms are designed for single bag query. In batch mode AL contexts the oracle is ask to label a set of query. The proposed algorithms do not implement a mechanism that ensure that bags contained in a set of query are different, which might be sub-optimal in this context.

\section{Conclusion}

This paper introduces two methods for MIAL in instance classification problems. Experiments show that leveraging the bag-level structure of data provides a significant reduction in the number of queries needed to accurate classifiers for difference benchmark problems. Future research includes studying how different types of structure and correlation within and between bags affect the behavior of MIAL algorithms. An extension of the methods should be proposed mitigate the effect of similar instance in a same bag and to improve the batch mode learning process. Finally, experiments will be conducted to measure the benefit of using MIAL on data collected from large real-world clinical contexts.

\bibliography{MIL1.bib}

\end{document}